\documentclass[11pt]{article}

\usepackage{acl2020}

\usepackage{times}
\usepackage{latexsym}
\usepackage{graphicx}
\usepackage{caption}
\usepackage{subcaption}
\usepackage[most]{tcolorbox}

\usepackage{microtype}

\title{What GPT Knows About Who is Who}

\author{Xiaohan Yang,  Eduardo Peynetti, Vasco Meerman \and  Chris Tanner \\
Institute for Applied Computational Science \\ Harvard University \\
\texttt{xiaohan\_yang@g.harvard.edu}, \texttt{eduardo.peynetti@gmail.com}, \\
\texttt{vmeerman@g.harvard.edu}, \texttt{christanner@g.harvard.edu}}
 \date{}

\begin{document}

\maketitle
\begin{abstract}


Coreference resolution -- which is a crucial task for understanding discourse and language at large -- has yet to witness widespread benefits from large language models (LLMs). Moreover, coreference resolution systems largely rely on supervised labels, which are highly expensive and difficult to annotate, thus making it ripe for prompt engineering. In this paper, we introduce a QA-based prompt-engineering method and discern \textit{generative}, pre-trained LLMs' abilities and limitations toward the task of coreference resolution. Our experiments show that GPT-2 and GPT-Neo can return valid answers, but that their capabilities to identify coreferent mentions are limited and prompt-sensitive, leading to inconsistent results.

\end{abstract}


\section{Introduction}
Coreference resolution (CR) aims to identify and cluster all words (i.e., mentions) that refer to the same entity or event. Solving this task is essential for natural language understanding, as mismatched references will lead to bias. Recent improvements in CR have been incremental \citep{lee-etal-2017-end, 10.1162/tacl_a_00300, cattan2020streamlining}, compared to other NLP tasks that have demonstrated more real-world impact. One reason is the limited training corpora. For example, one of the primary datasets, ECB+ \citep{cybulska-vossen-2014-using}, contains only 984 documents, including 6,833 mentions and 2,741 clusters. Moreover, this dataset was built around 43 news topics ten years ago, potentially leading to generalization problems for the state-of-the-art (SOTA) models.

When dealing with low-resource tasks, there is an emerging trend to perform \textit{prompt engineering} with pre-trained LMs. Unlike fine-tuning \citep{brown2020language, wei2022finetuned}, prompt engineering does not update the pre-trained model's weights when completing the downstream task. Instead, one transforms the downstream task to match the original task of the pre-trained model \citep{Liu2021PretrainPA}. For example, for machine translation, one can create prompts such as ``English: I love bread. French:'' and input them to a generative LM (e.g., GPT-2). If the pre-trained model encountered similar patterns during training, it should be able to generate the translated French sentence. Nevertheless, to the best of our knowledge, there is scarce research on applying this approach to coreference resolution \citep{sanh2021multitask}.

To better understand if pre-trained LMs can help resolve coreferences, we construct a QA-based prompting method and experiment with both GPT-2 \citep{radford2019language} and GPT-Neo \citep{gao2020pile}. By using this prompting methodology, we measure if these models can predict whether two mentions are coreferent. For evaluation, we use the ECB+ dataset, which provides gold mentions and clustering labels. We compare the results with unsupervised and supervised coreference resolution models, including a classic rule-based system \citep{lee-etal-2011-stanfords}, the seminal end-to-end neural model \citep{lee-etal-2017-end}, and a recent SOTA model \citep{cattan2020streamlining}.

\begin{figure*}[ht]
  \includegraphics[width=\textwidth,height=4cm]{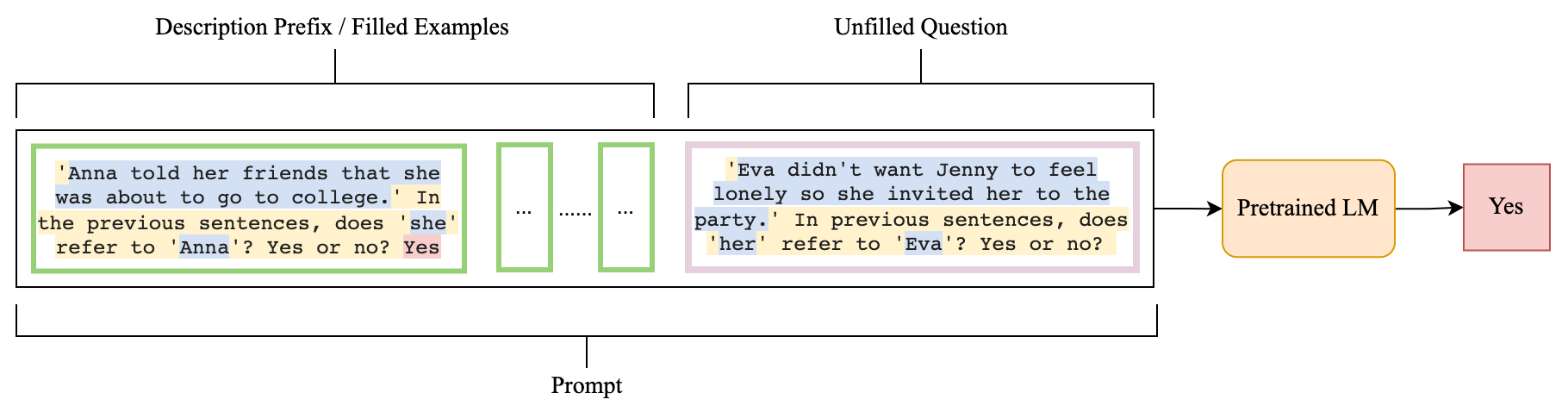}
  \caption{An example of prompt-based learning for CR. The green block represents the prefix, which serves as the description of the CR task and remains unchanged throughout an experiment for all inputs $x$. The purple block is the unfilled prompt, which changes for each input $x$ and serves as the prediction. Moreover, in each block, the yellow part is the prompting function while the blue and red parts are the original data $x$ and $y$, respectively.}
  \label{prompt_example}
\end{figure*}

\section{Related Work}
\paragraph{Prompt-based learning}

Prompt-based learning is a fast-growing area in NLP, as it can reduce the need to fine-tune models and rely on supervised labels. According to the survey by \citeauthor{Liu2021PretrainPA}, over 120 papers have been published since 2019, which collectively demonstrates effectiveness toward many different tasks: text classification \citep{tam-etal-2021-improving,Holtzman2021SurfaceFC}, factual probing \citep{perez2021true}, question-answering \citep{tsimpoukelli2021multimodal}, and more. Nevertheless, to the best of our knowledge, only one prompt-based learning paper concerned CR. Specifically, \citeauthor{sanh2021multitask} introduces T0, a zero-shot generalization of T5 \citep{JMLR:v21:20-074}. The authors convert various supervised datasets into task-specific prompts, including CR. Using the WSC dataset \citep{levesque2012winograd}, they achieve over 60\% accuracy. Although this result is not comparable with supervised state-of-the-art (SOTA) models, it still offers compelling results and suggests the model might contain CR knowledge without requiring supervised training on the task. However, since the WSC dataset only focuses on highly ambiguous pronouns, it is not as complete as the standard CR task that involves named and nominal mentions.

\paragraph{Traditional CR Models}

Similar to other NLP tasks, most CR models can be categorized as being either unsupervised or supervised. A commonly used unsupervised model is the \textit{Multi-Pass Sieve} model \citep{lee-etal-2011-stanfords}. This rule-based system extracts entity mentions and clusters them by applying 13 ``filters'' in successive manner. Amongst supervised models, \textit{e2e-coref} \citep{lee-etal-2017-end} is the seminal end-to-end neural model. This model performs within-document CR and was trained on the OntoNotes (CoNLL-2012) dataset. Building on this architecture, \citet{cattan2020streamlining} performs cross-document CR for entities and events by training on the ECB+ dataset and using RoBERTa \citep{liu2019roberta} as an encoder. Although supervised models offer significant improvements over unsupervised models, they are expensive to train; most SOTA models have $O(n^4)$ complexity, where $n$ is the length of each document.

\section{Methodology}

This section introduces our prompt-based learning method for CR. Typically, CR models can be broken down into three sub-tasks: (1) detecting mentions; (2) predicting whether two given mentions are coreferent or not; (3) and clustering mentions accordingly. The crux of CR research centers around the second part, which is also our focus.


Building on the approach introduced by \citet{sanh2021multitask}, we define our input $x$ as $[text, m_1, m_2]$ and output $y$ as a binary label. Specifically, $m_1$ and $m_2$ are a pair of gold mentions in a document, and the $text$ are the sentences containing those mentions. For example, in Figure \ref{prompt_example}, within each green box, the successive blue parts are $text$, $m_1$, $m_2$, respectively. We define a prompting function $f$, which takes $x$ as input and produces a question prompt $q_x$ (Equation \ref{eq: fprompt}). Further details about $f$ are in Appendix \ref{sec: prompt_formulas}.

\begin{equation} \label{eq: fprompt}
q_x = f(x) 
\end{equation}

Moreover, to allow the model to understand the task, we use few-shot learning \citep{triantafillou2017fewshot} by constructing a filled prefix. In particular, we select $k$ examples, $A$, from the training dataset and feed these examples into the same prompting function $f$. Then, we append the true label (`Yes' or `No') to the outputs, yielding the filled prefix $q_A$ (Equation \ref{eq: kprompts}). To be clear, each individual prefix $q_{i \in k}$ constitutes a single green box in Figure \ref{prompt_example}.
\begin{equation} \label{eq: kprompts}
q_A = f(A) 
\end{equation}

Last, adding the unfilled prompt $q_x$ to the filled prefix $q_A$ will give us the full prompt for data point $x$. This allows us to get a prediction $z$ without updating any parameters $\theta$ in the pre-trained LM.
\begin{equation} \label{eq: model}
z = P(q_A+q_x;\theta) \\
\end{equation}

Since we use pre-trained LMs directly, without fine-tuning, we do not have control over its output; the model can generate invalid answers beyond our desired outputs, `Yes' or `No'. Therefore, we repeat the process $m$ times to get a more robust prediction $\bar z $. To mitigate the bias of one specific $f$, we average the output of $n$ different prompt formulas to get the final prediction (Equation \ref{eq: aggf}).

\begin{equation} \label{eq: aggf}
y =  \frac{\sum_{i=1}^n \bar z_i}{n}
\end{equation}


\section{Experimental Setup}
\label{sec: setup}

\paragraph{Datasets} 
We use the ECB+ dataset \citep{cybulska-vossen-2014-using} as our input source, which contains both within- and cross-document coreference information for both event and entity mentions. This dataset consists of 984 documents around 43 news topics, among which 196 documents are in the development set. After preprocessing the data, as described in Appendix \ref{sec: preprocessing}, our development set consists of 172 documents.  

To generate a prefix $x_0$, we experiment with three data sources: the training sets of WSC \citep{levesque2012winograd} and ECB+ \citep{cybulska-vossen-2014-using}, and a simple dataset that we manually generated. The WSC dataset was used in the research most similar to ours, T0 \citep{sanh2021multitask}, which we compare against while using much smaller pretrained LMs (i.e., GPT-2 and GPT-Neo). As mentioned, ECB+ provides more natural and comprehensive references than WSC. Our manually generated dataset uses 10 very simple examples -- allowing one to discern the impact on performance.


When using the ECB+ dataset, we only considered pairs of mentions that are within the same or successive sentences. When evaluating our model, we considered all mention-pair combinations, $[m_1,m_2]$, within said sentences. Relying on the gold mentions, we obtain a dataset with $17832$ candidate mention pairs, among which $7.86\%$ are positive samples. Finally, we apply 5 prompt functions from \citeauthor{sanh2021multitask} to generate the full prompts.  


\paragraph{Models}
We used three traditional CR models as baselines: \textit{Multi-Pass Sieve} \citep{lee-etal-2011-stanfords}, the seminal end-to-end neural model (\textit{e2e-coref}) \citep{lee-etal-2017-end}, and a SOTA extension (the \textit{Streamlining} model) \citep{cattan2020streamlining}. Respectively, these models represent three categories: a rules-based model, a supervised model trained on a different dataset, and a supervised model trained on the same dataset. In terms of implementations, we use the CoreNLP toolkit for the \textit{Multi-Pass Sieve} model \citep{manning2014stanford} and AllenNLP \citep{DBLP:journals/corr/abs-1803-07640} for \textit{e2e-coref}. Since there is no publicly available pre-trained \textit{Streamlining} model \citep{cattan2020streamlining}, we fully train the model from scratch using a V100 GPU on Google Colab. To fairly compare with other models, we set a $0.5$ threshold for the pairwise scorer in the Streamlining model. We evaluate all models by their mention pairwise scorers, not their clustering decisions.

Limited by our computational resources, we choose GPT-2 and GPT-Neo-125M as our pre-trained LMs \footnote{Our code can be found at \url{https://github.com/AwesomeCoref/prompt-coref}}. During inference, the output token length is set to 1, since our expected output is one word (i.e., `Yes' or `No'). To generate more robust results, the repetition parameter $m$ is set to $5$. We ran our text generative models with multiple temperature settings ranging from 0 to 1, none of which produced significant changes. We settled on using a value of $0.7$, to limit the greediness of the generated responses. In terms of few-shot learning, we experimented with $k \in \{0,2,4,10\}$ and display the results from the 4-shot setting since it produces the best accuracy. To reduce bias introduced by prefixes, we ensure each prefix has equally-balanced samples. For example, for the 4-shot setting, the filled prefix will have 2 positive examples and 2 negative examples. 

\section{Results and Analysis}

\begin{table}[ht]
\centering
\begin{tabular}{lrr}
\hline & \textbf{Yes/No Predictions} \\ \hline
0-shot & 5\% \\
2-shot & 93.7\% \\
4-shot & 96.2\% \\
10-shot & 98\%  \\
\hline
\end{tabular}
\caption{\label{valid-pred} Percentage of Yes/No predictions by GPT-2}
\end{table}
We first question if GPT-based models can produce valid answers. In Figure \ref{valid-pred}, we observe that GPT-2 predicts `Yes' or `No' for over $93.7\%$ samples when at least 2 filled prefixes are provided.

\begin{figure}[ht]
\includegraphics[width=\linewidth]{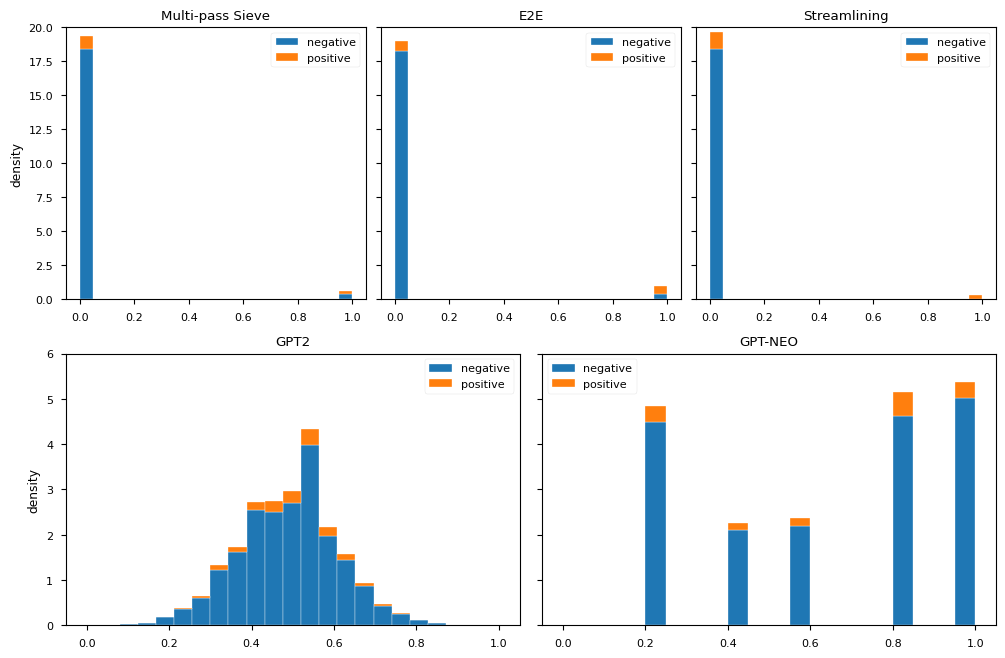}
\caption{Distribution of predicted values }
  \label{fig:pred_distribution}
\end{figure}

However, although the answers are valid, they are inaccurate. In Figure \ref{fig:pred_distribution}, we plot the distribution of predicted labels for each model, where the red bars denote the distribution of positive examples (ground truth), and the blue bars denote negative ones (ground truth). Traditional CR models generally predict low values for negative examples, indicated by blue bars being concentrated at $0$. As for positive examples, \textit{e2e-coref} shows better precision since more positive examples are classified correctly at $1$. Yet, GPT-2 seems to be both sensitive to prompts and unstable over the repetitions of each prompt. Furthermore, GPT-Neo's predictions are inaccurate and no better than random, even though it predicts consistent results for multiple runs with the same prompt. 

Similar conclusions can be drawn from Table \ref{overall-performance}, where GPT-based models have the lowest AUC and F1 scores. Specifically, the extremely low precision causes the bad results. Since the ECB+ dataset is highly imbalanced, random predictions from GPT-based models will lead to a low precision, reflecting the proportion of positive samples. For completeness, we also perform an experiment on the WSC dataset (see GPT-2$_{wsc}$), which is a test dataset used by \citet{sanh2021multitask}. GPT-2 also fails on this task, as its mean prediction averaged across different prompts is always ``Yes'' .


\begin{table}[ht]
\centering
\begin{tabular}{lrrrrr}
\hline & \textbf{Acc} & \textbf{Prec} & \textbf{Rec} & \textbf{F1} & \textbf{AUC}\\ \hline
Multi Sieve  & 0.93 &       0.39 &    0.20 & 0.27  & 0.59\\
e2e-coref  & 0.95 &  0.62 & 0.46 & 0.53  & 0.72\\
Streamline  & 0.93 &  0.87 & 0.19 & 0.31 & 0.59 \\
GPT-2 & 0.50 & 0.08 & 0.53 & 0.14 & 0.51 \\
GPT-NEO & 0.38 & 0.08 & 0.68 & 0.15 & 0.52 \\
GPT-2$_{wsc}$ & 0.37 &       0.37 &    1.00 & 0.54 & 0.50 \\
\hline
\end{tabular}
\caption{\label{overall-performance}  Performance of different models.}
\end{table}

\begin{figure}[ht]
\includegraphics[width=\linewidth]{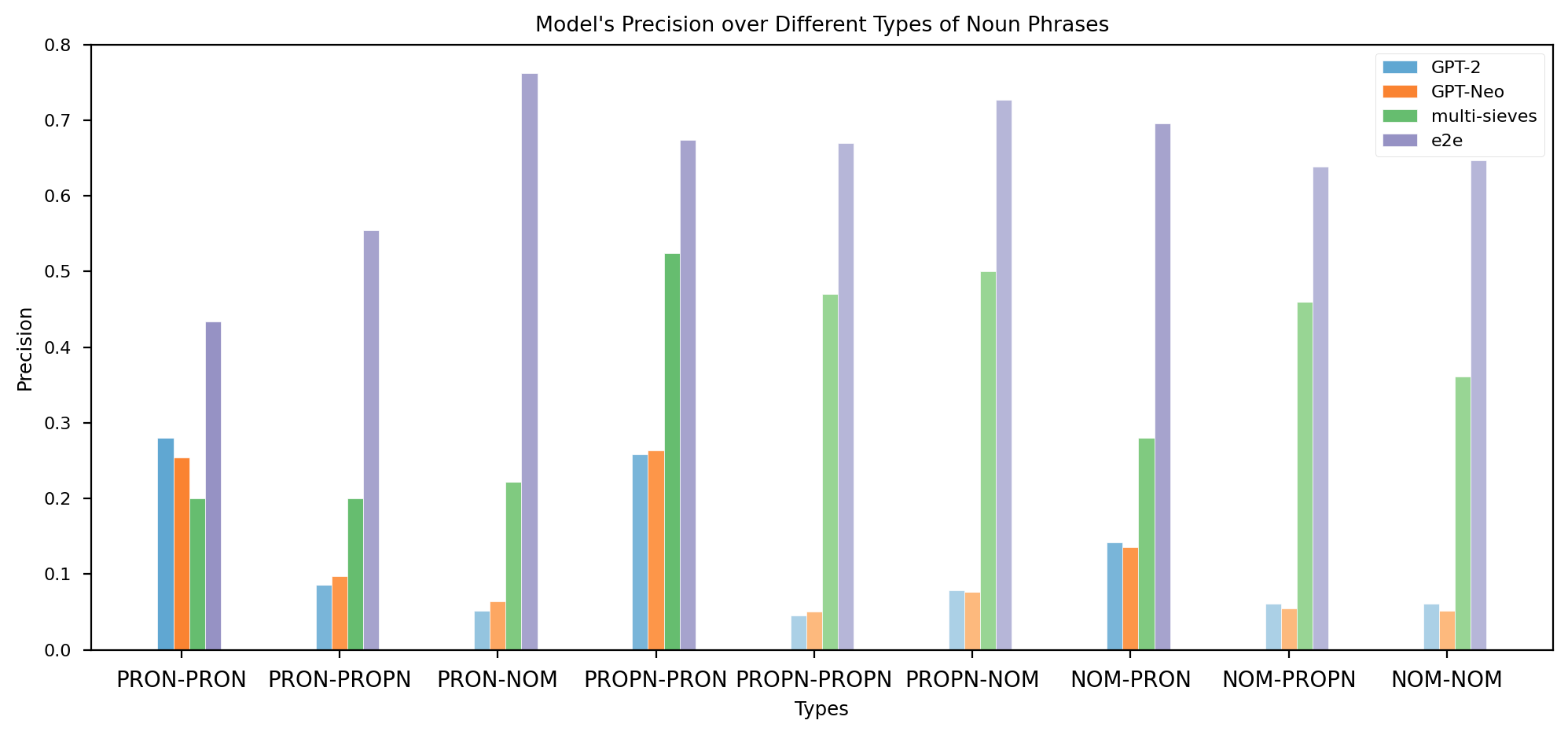}
\caption{Model's precision over various types of noun phrases, including pronouns, proper nouns and nominal nouns. Each bar's hue intensity denotes the data density.}
\label{fig:pos}
\end{figure}

\paragraph{POS and Entity Types}
While the overall performance indicates that GPT models are comparable to a random model, we hypothesize that for some subset of mention pairs, GPT might perform better. To investigate, we conducted experiments based on part-of-speech (POS) tags and named-entity types. Figure \ref{fig:pos} shows that both GPT-2 and GPT-Neo can capture coreferent relationships relatively better when the second mention is a pronoun. Moreover, this trend is stronger when the first mention is a pronoun or a proper noun. Nonetheless, \textit{e2e-coref} performs better than both GPT models across all POS tags, and the gap is widest when the second mention is a nominal noun phrase.

\begin{figure}[ht]
\centering
\begin{subfigure}{0.5\linewidth}
  \centering
  \includegraphics[width=\linewidth]{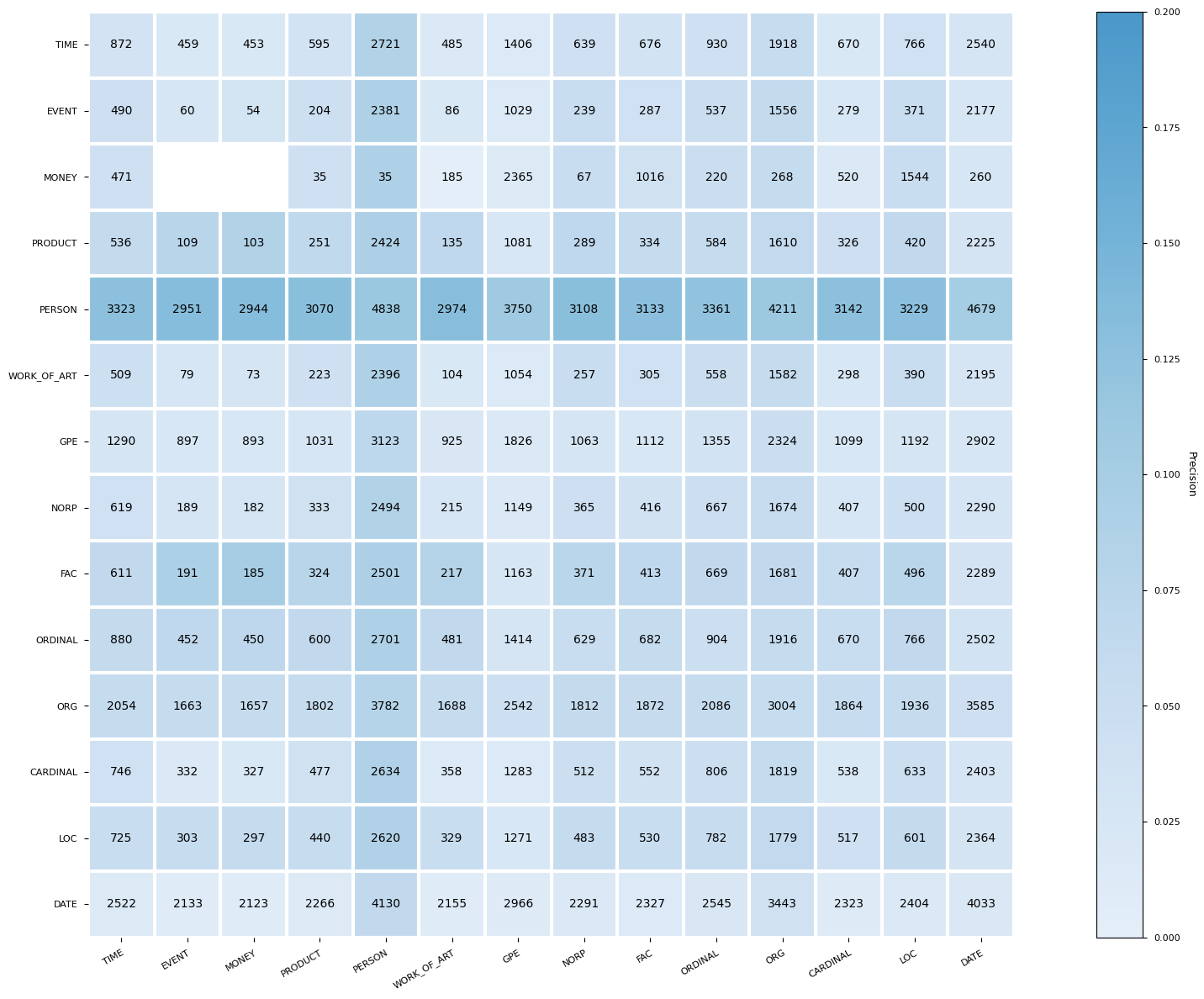}
  \caption{GPT-2}
  \label{fig:sub2A}
\end{subfigure}%
\begin{subfigure}{0.5\linewidth}
  \centering
  \includegraphics[width=\linewidth]{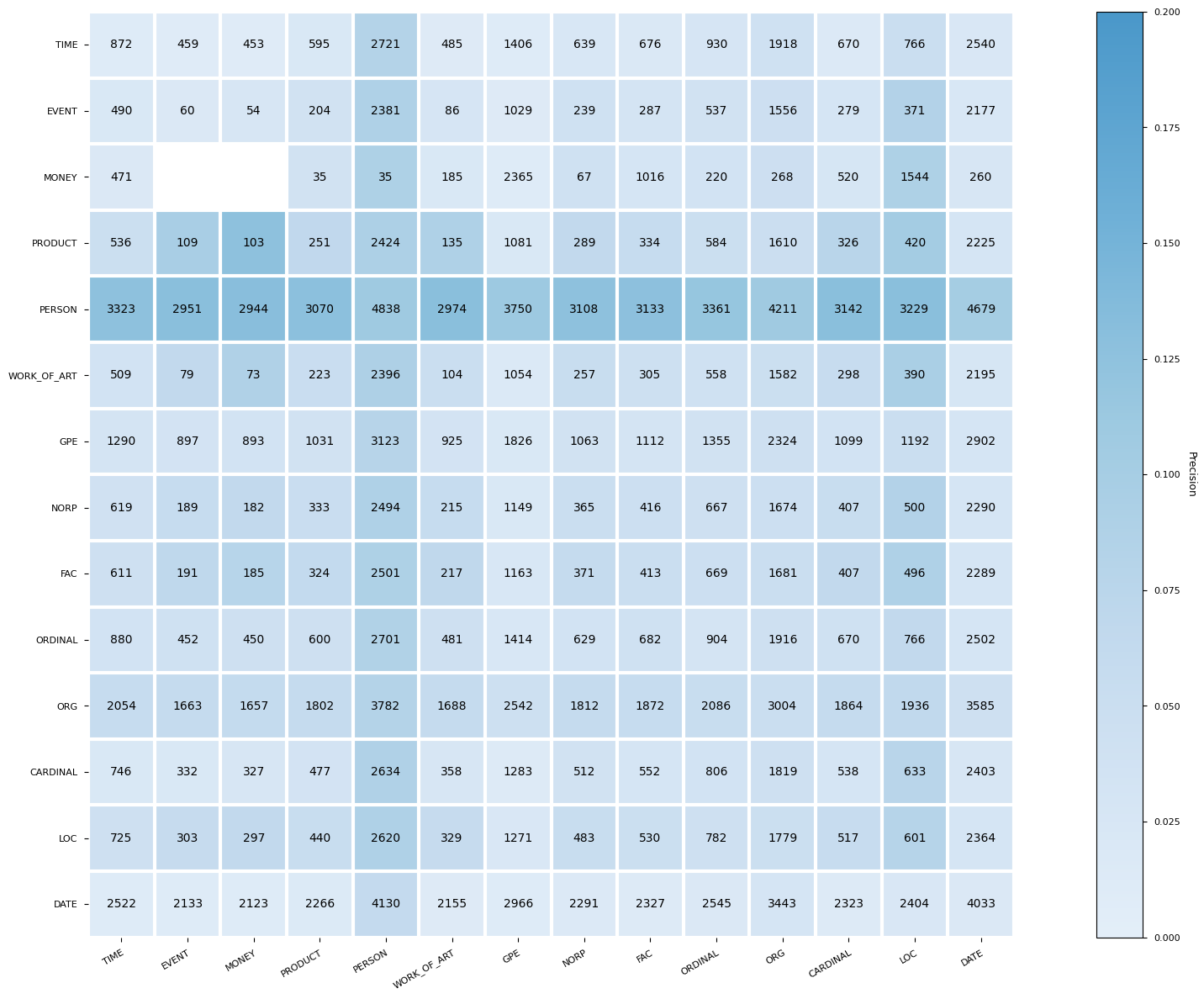}
  \caption{GPT-Neo}
\end{subfigure}

\begin{subfigure}{0.5\linewidth}
  \centering
  \includegraphics[width=\linewidth]{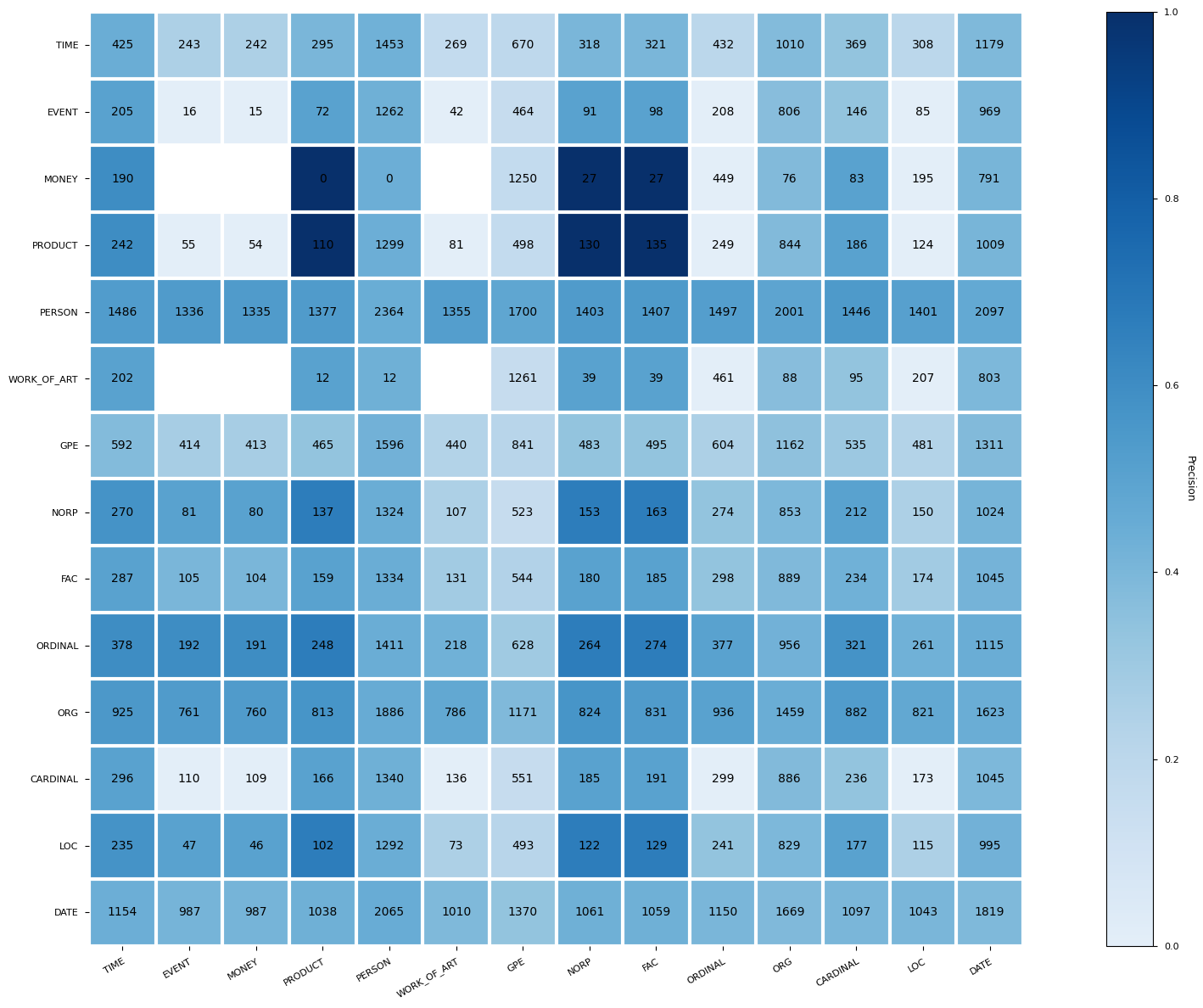}
  \caption{Multi-pass Sieve}
  \label{fig:sub1A}
\end{subfigure}%
\begin{subfigure}{0.5\linewidth}
  \centering
  \includegraphics[width=\linewidth]{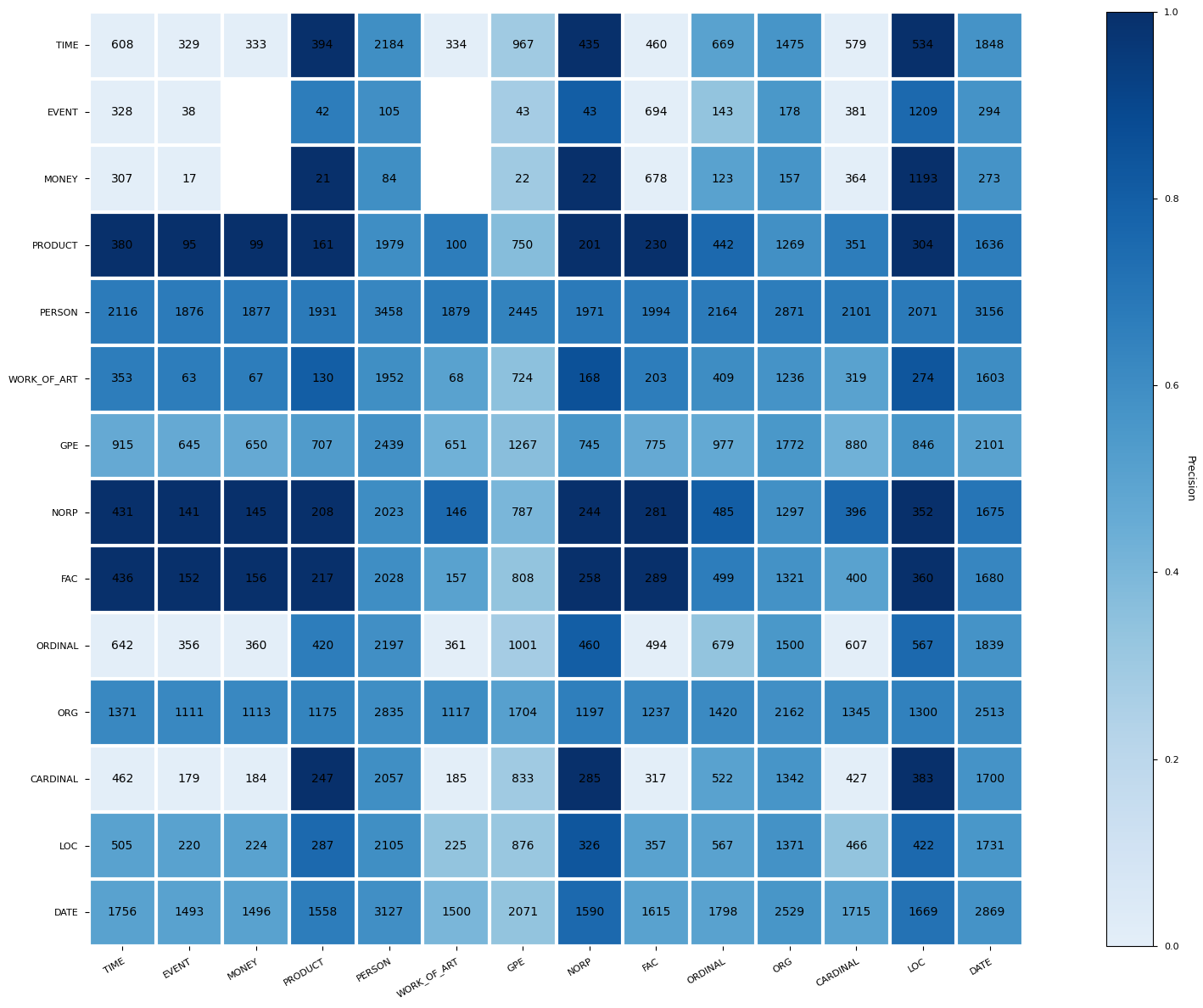}
  \caption{e2e-coref}
\end{subfigure}
\caption{GPT-2's performance on different named-entity types. We use colors to denote performance and the text to show data density in each category. 
}
\label{fig:named entity}
\end{figure}

As for named entities, Figure \ref{fig:named entity} shows that both GPT-2 and GPT-Neo perform better in precision when one mention is of type PERSON. Moreover, GPT-Neo can identify coreferent relationships more precisely if the second mention is Non-GPE locations (i.e., LOC).
However, their precision scores are far lower than the scores from classical CR models. In particular, both the multi-pass sieve model and e2e-coref model reach the highest precision when a mention is a PRODUCT object (e.g., vehicle, food) or a NORP object (e.g., nationality, religious or political group).   


\begin{figure}[ht]
  \includegraphics[width=\linewidth]{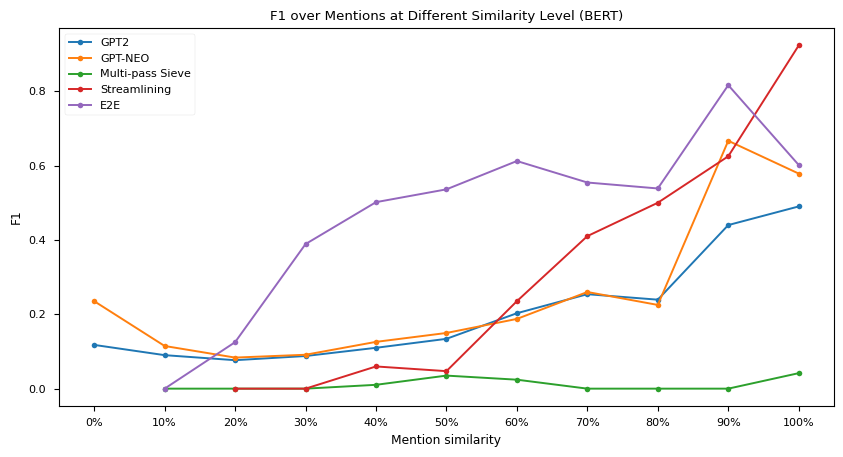}
  \caption{Different models' F1 score over various level of mention similarities based on BERT embedding.}
  \label{fig:Mention Similarity}
\end{figure}

\paragraph{Mention Similarity}  
In addition to inspecting how performance varies with mention \textit{types}, we also considered how performance is affected by mentions' similarity. Using pre-trained BERT \citep{devlin-etal-2019-bert}, we encode each mention into span representations by averaging its tokens' last hidden states. Then, we measure cosine similarity between mention pairs. 

Figure \ref{fig:Mention Similarity} shows that F1 scores generally improve as the semantic similarity increases. Although, the multi-pass sieve model maintains a low F1 because it is a rule-based model that tends to predict False for most samples -- which yields a high accuracy for unbalanced datasets. The e2e-coref model performs well on mentions that are not so similar, while the performance of Streamlining model improves drastically as similarity is greater than \text{50\%}. However, both GPT-2 and GPT-NEO yield low F1 (approximately \text{0.2}) for mention pairs with less than \text{70\%} similarity. When considering mentions of higher similarity, GPT-based models can achieve over 0.4 F1 score. 


\section{Conclusion}
In this paper, we rely on prompt-based learning to analyze how much GPT-like models know about coreference resolution. Despite the popularity of prompting in recent NLP research, we find that LLMs perform poorly on this task without fine-tuning. Nonetheless, these models achieve relatively better performance for specific types of mentions, including pronouns and person objects, and mention pairs with high similarity.

\bibliography{annotate_ref}
\bibliographystyle{acl_natbib}

\appendix

 \newpage

\section{Prompt formulas}

\label{sec: prompt_formulas}

\begin{figure}[ht]
  \includegraphics[width=\linewidth]{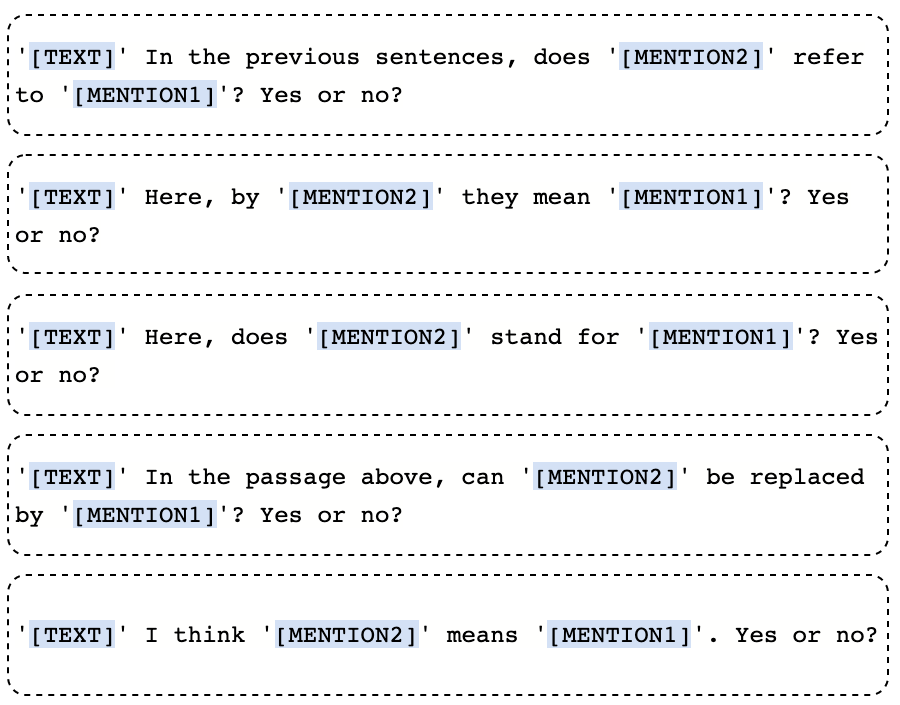}
  \caption{Prompt Formulas. We experiment with these 5 prompt formulas mentioned in \citet{sanh2021multitask}. Here, each block is one formula and the parts highlighted in blue are $[text, m_1, m_2]$ respectively. }
  \label{fig:prompt_formulas}
\end{figure}

\section{Data Preprocessing}
\label{sec: preprocessing}

The original ECB+ dataset is in XML format, where everything is tokenized. Moreover, the information about gold mentions and gold clusters is related to token ids. However, we cannot easily get the plain text by joining tokens with a space character. If we do so, we will get strange looking text as shown below.  \\

\noindent\fbox{\begin{minipage}{\linewidth}
http : / / www . accesshollywood . com / lindsay - lohan - leaves - betty - ford - checks - into - malibu - rehab \_ article \_ 80744 Lindsay Lohan Leaves Betty Ford , Checks Into Malibu Rehab First Published : June 13 , 2013 4 : 59 PM EDT Lindsay Lohan has left the Betty Ford Center and is moving to a rehab facility in Malibu , Calif . , Access Hollywood has confirmed .

\end{minipage}} \\

In this example, we can see objects like urls, datetime and punctuation are not in the right format. Since we are using the text as an input to the prompt function, we need to properly format them to align with normal text that GPTs are trained on. Moreover, as gold mention and gold clusters are based on original token ids in ECB+, when we parsed and re-formatted the data, we could match these ids again. Continuing with the previous example, our parsing algorithm cleans up the previous text to be something as follows.\\

\noindent\fbox{\begin{minipage}{\linewidth}
http://www.accesshollywood.com/lindsay-lohan-leaves-betty-ford-checks-into-malibu-rehab\_article\_80744 [EOS] Lindsay Lohan Leaves Betty Ford, Checks Into Malibu Rehab First Published: June 13, 2013 4: 59 PM EDT [EOS] Lindsay Lohan has left the Betty Ford Center and is moving to a rehab facility in Malibu, Calif., Access Hollywood has confirmed. [EOS]
\end{minipage}} \\

\section{Additional Results}
\label{sec: add_results}
Here are additional results for our experiments. 

\paragraph{Experiments on Prefix}
The aggregate results from few shot learning are displayed in Table \ref{n-shot}.
Our results show that 4-shots learning performs the best for both GPT-2 and GPT-NEO in terms of accuracy. Unexpectedly, as we increase the size of examples, the result does not improve accordingly. Given 10 examples in prefix, the model tend to predict ``yes'' more easily. One possible explanation might be that we have balanced examples in prefix while the actual querying data only have around 8\% positive samples.  

\begin{table}[ht]
\centering
\begin{tabular}{lrrrrr}
\hline & \textbf{Acc} & \textbf{Prec} & \textbf{Recall} & \textbf{F1} & \textbf{AUC} \\ \hline
2-shot & 0.39 & 0.08 & 0.64 & 0.14 & 0.50 \\
4-shot & 0.51 & 0.08 & 0.51 & 0.14 & 0.51 \\
10-shot & 0.19 & 0.08 & 0.90 & 0.15 & 0.51 \\
\hline
\end{tabular}
\caption{\label{n-shot} n-shot performance from the text generative models}
\end{table}

\begin{table}[ht]
\centering
\begin{tabular}{lrrrrr}
\hline & \textbf{Acc} & \textbf{Prec} & \textbf{Recall} & \textbf{F1} & \textbf{AUC} \\ \hline
simple    & 0.61 &       0.08 &    0.36 & 0.13 & 0.50 \\
WSC & 0.08 &       0.08 &    1.00 & 0.15 & 0.50 \\
ECB+       & 0.54 &       0.08 &    0.48 & 0.14 & 0.51 \\
\hline
\end{tabular}
\caption{\label{prefix_data} Average results from each dataset that is used for the experiments}
\end{table}

Moreover, we experiment with various datasets for prefix as discussed in section \ref{sec: setup}. The results in Table \ref{prefix_data} shows that prefix does have an impact on the results. The prefix generated from ECB+ dataset performs slightly better than others regarding to AUC. This is understandable because we evaluate on the ECB+ development set. Beyond our expectation, WSC-prefix result in a perfect recall and a super bad accuracy, which means that this prefix lead models to generate ``yes'' regardless of the context. This result further proves that GPT-2 is very sensitive to prompts.



\end{document}